\def\year{2020}\relax
\documentclass[letterpaper]{article} 
\usepackage{aaai20}  
\usepackage{times}  
\usepackage{helvet} 
\usepackage{courier}  
\usepackage[hyphens]{url}  
\usepackage{graphicx} 
\usepackage{xcolor}
\usepackage{amsmath}
\usepackage{adjustbox}
\usepackage{multirow}
\usepackage{array}
\usepackage{url}

\newcolumntype{x}[1]{>{\centering\arraybackslash\hspace{0pt}}p{#1}}

\newcommand{\ie}{\textit{i}.\textit{e}., }
\newcommand{\eg}{\textit{e}.\textit{g}., }

\title{Scale-wise Convolution for Image Restoration}
\author{
Yuchen Fan\textsuperscript{\rm 1},
Jiahui Yu\textsuperscript{\rm 1},
Ding Liu\textsuperscript{\rm 2},
Thomas S. Huang\textsuperscript{\rm 1}\\
\textsuperscript{\rm 1}University of Illinois at Urbana-Champaign,
\textsuperscript{\rm 2}Bytedance Inc.\\
yuchenf4@illinois.edu, jyu79@illinois.edu, liuding@bytedance.com, t-huang1@illinois.edu
}
\begin{document}

\maketitle

\begin{abstract}
While scale-invariant modeling has substantially boosted the performance of visual recognition tasks, it remains largely under-explored in deep networks based image restoration. Naively applying those scale-invariant techniques (\eg multi-scale testing, random-scale data augmentation) to image restoration tasks usually leads to inferior performance. In this paper, we show that properly modeling scale-invariance into neural networks can bring significant benefits to image restoration performance. Inspired from spatial-wise convolution for shift-invariance, ``scale-wise convolution'' is proposed to convolve across multiple scales for scale-invariance. In our scale-wise convolutional network (SCN), we first map the input image to the feature space and then build a feature pyramid representation via bi-linear down-scaling progressively. The feature pyramid is then passed to a residual network with scale-wise convolutions. The proposed scale-wise convolution learns to dynamically activate and aggregate features from different input scales in each residual building block, in order to exploit contextual information on multiple scales. In experiments, we compare the restoration accuracy and parameter efficiency among our model and many different variants of multi-scale neural networks. The proposed network with scale-wise convolution achieves superior performance in multiple image restoration tasks including image super-resolution, image denoising and image compression artifacts removal. Code and models are available at: \url{https://github.com/ychfan/scn_sr}.
\end{abstract}

\section{Introduction}

The exploitation of scale-invariance in computer vision and image processing has greatly benefited feature engineering~\cite{lowe2004distinctive}, image classification~\cite{szegedy2015going}, object detection~\cite{cai2016unified,lin2017feature,fu2017dssd,singh2018analysis,yu2016unitbox}, semantic segmentation~\cite{ronneberger2015u,badrinarayanan2017segnet} and the training of convolutional neural networks~\cite{szegedy2015going,singh2018sniper}. For example, Lowe proposed the scale-invariant features (SIFT)~\cite{lowe2004distinctive} that are consistent across a substantial range of affine distortion. Due to its robustness, the SIFT features have been shown undeniably successful and ubiquitously used for image matching, image classification, object detection and many others computer vision tasks. More recently, the importance of scale-invariance is demonstrated in object detection using convolutional neural networks~\cite{singh2018analysis}, which is further improved by an efficient multi-scale training method~\cite{singh2018sniper} of object detectors.

While modeling scale-invariance has greatly improved visual recognition performance, few efforts have succeeded in utilizing scale-invariant features for image restoration tasks based on convolutional neural networks. For example, random-scale data augmentation during training and multi-scale testing are proven beneficial to a wide range of image recognition problems like classification~\cite{szegedy2015going} and detection~\cite{fu2017dssd}. However, naively applying those techniques usually lead to worse performance for image restoration like super-resolution. It is likely due to the fact that, unlike image recognition problems, the performance of low-level image restoration tasks is very sensitive to naive scaling of input images. Is modeling scale-invariance unnecessary for image restoration tasks?

In this work, we demonstrate that properly modeling scale-invariance into convolutional neural networks can bring significant gain for image restoration tasks. We draw inspiration from spatial-wise convolution for shift-invariance, and propose the ``scale-wise convolution'' that is designed to convolve across multiple scales to achieve scale-invariance. In our scale-wise convolutional network (SCN), we first map the input image to the feature space and then build a feature pyramid representation via bi-linear down-scaling progressively. Based on the feature pyramid, we apply the same function (several shared layers) to process different scales and aggregate the neighborhoods. After several residual blocks with scale-wise convolutions, we take the feature output of the largest scale and predict the results with a linear convolution. The proposed scale-wise convolution learns to dynamically activate and aggregate features from different input scales in each residual building block, in order to exploit contextual information on multiple scales.

We compare our proposed network to several multi-scale convolutional networks with different design choices (shared or un-shared weights, different layer connectivity, etc). We show in our experiments that these multi-scale convolutional networks are inferior than our proposed scale-wise convolutional network. We also study the way of feature fusing across different scales: \ie(1) convolution and deconvolution with stride, (2) average pooling down-sampling and nearest neighbour up-sampling, and (3) bilinear re-sampling (both down-sampling and up-sampling). We further study the number of scales and the scale factor in our scale-wise convolutional networks and analyze their performances.

To demonstrate the effectiveness of our proposed scale-wise convolutional networks, we conduct extensive experiments on image super-resolution, denoising and compression artifacts removal. Our experiments reveal that: (1) Multi-scale models significantly improve the performance compared with single-scale models. (2) Scale-wise convolution for cross-scale modeling is more parameter-efficient than general multi-scale models, and achieves better efficiency-accuracy trade-offs. (3) Scale-wise convolution benefits from more scales in feature pyramid and proper scaling ratios in-between. Moreover, our proposed SCN can achieve superior results to state-of-the-art methods with even fewer parameters on image super-resolution, denoising and compression artifacts removal.


\section{Related Work}

\subsection{Scale-Invariant Modeling in Visual Perception}
The prior of scale-invariance in image processing has motivated many deep neural network architectures and models. For example, a scale-invariant convolutional neural network (SiCNN) proposed by~\cite{xu2014scale} is designed to incorporate multi-scale feature exaction and classification into the multi-column architecture. These columns share a same set of weights which are transformed for different scales. SiCNN improves the performance of image classification by learning the feature in different scales in different columns.  Multi-scale architectures like U-Net~\cite{ronneberger2015u} and SegNet~\cite{badrinarayanan2017segnet} incorporate multiple scales in different network staged and inter-connects them, which are proven to have higher performance on image segmentation tasks. Feature Pyramid Network~\cite{lin2017feature} proposed a top-down architecture with lateral connections for building high-level semantic feature maps at all scales. Moreover, during the training and testing of deep neural networks, multiple scales can also be useful for a wide range of applications like classification and detection. 

\subsection{Multi-Scale Architectures in Image Restoration} 
Several approaches have explored involving multiple scales into neural network architecture for image restoration tasks. The Dual-State Recurrent Network (DSRN)~\cite{han2018image} proposed a neural network to jointly utilize signals on both low-resolution scale and high-resolution scale for image super-resolution. Specifically, recurrent signals in DSRN are exchanged between these two scales in both directions via delayed feedback. The Deep Back-Projection Networks~\cite{haris2018deep} exploited multiple iterative up-sampling and down-sampling layers, which provides an error feedback mechanism for projection errors at each stage of image super-resolution. Balanced two-stage network~\cite{fan2017balanced} is decoupled into scales along layers. Multi-scale residual network (MSRN) is proposed to fully exploit the image features~\cite{li2018multi}, in which the convolution kernels of different sizes can adaptively detect the image features in different scales.

\subsection{Weight Sharing in Image Restoration}
For image restoration tasks, the receptive field of a convolutional network has a critical role as it determines the amount of contextual information that can be used for prediction. However, naively increasing the number of layers leads to computationally expensive models with low parameter efficiency. To make a model compact, recursive neural networks~\cite{kim2016deeply,tai2017image} are proposed by sharing weights repeatedly among different recurrent stages. The recurrent architecture has also been used in ~\cite{han2018image,liu2018non}.


\section{Scale-wise Convolutional Networks}
\label{sec:method}

In this section, we introduce our approach in the following manner. We first describe the scale-wise convolution operator and discuss its properties in details. Then the scale-wise convolutional networks are developed based on repeatedly constructing residual blocks with scale-wise convolution. Finally we discuss a number of design choices and construct several multi-scale network variants as our baselines. 

\subsection{Scale-wise Convolution}
\begin{figure}
  \centering
  \includegraphics[width=1.0\columnwidth]{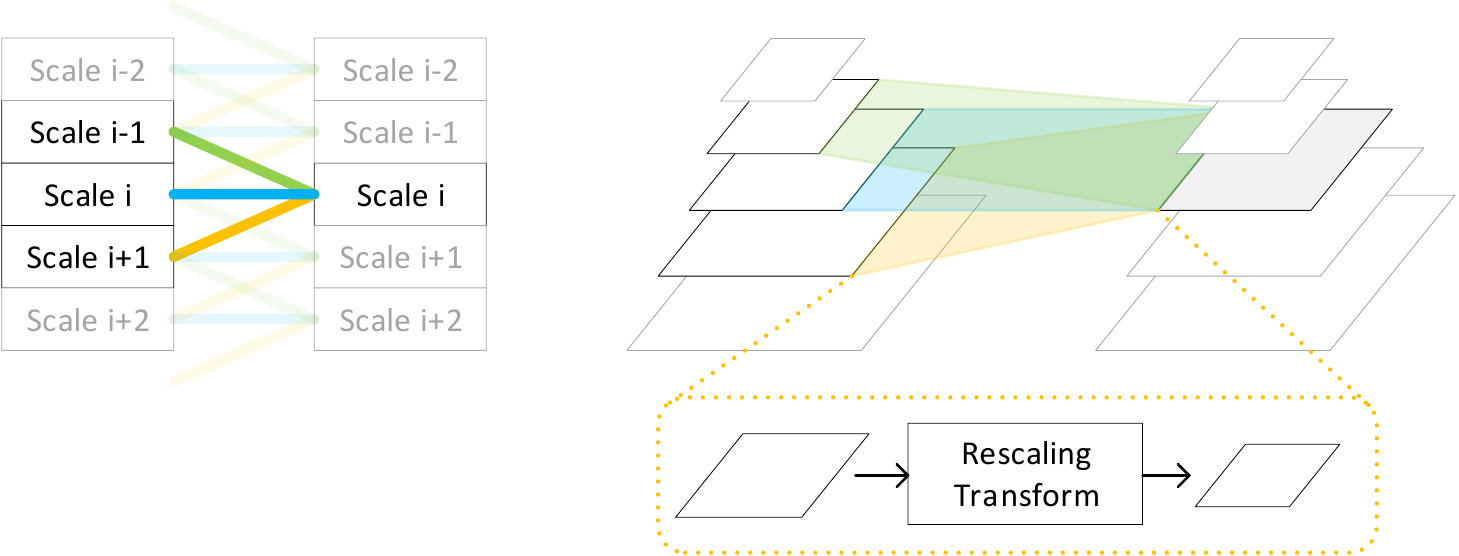}
  \caption{Illustration of scale-wise convolution. Left: projected view of scale-wise convolution over features in different scales. Right: scale-wise convolution as sum of features from neighbour scales transformed by $f_i$.}
  \label{fig:scale_conv}
\end{figure}

We propose the scale-wise convolution operator that convolves features along the scale dimension from a multi-scale feature pyramid. Figure \ref{fig:scale_conv} visually illustrates the idea of ``convolution'' across scales.

Formally, the scale-wise convolution $\mathcal{F}$ takes a multi-scale feature pyramid $\boldsymbol{X}^l = \{\boldsymbol{x}_s^l\}_{s=1}^N$ from layer $l$ with multiple scales from $1$ to $N$, and generates the next feature pyramid with the same size  $\boldsymbol{X}^{l+1}=\mathcal{F}(\boldsymbol{X}^l)$ by ``convolution'' across scales. The output feature $\boldsymbol{x}_s^{l+1}$ is transformed from features of previous neighbouring scales $\{\boldsymbol{x}_{s-k}^{l}, \dots, \boldsymbol{x}_{s}^{l}, \dots, \boldsymbol{x}_{s+k}^{l}\}$. Specifically, the output of scale-wise convolution with kernel size $2k+1$ for layer $l+1$ and scale $s$ is computed as

$$
\boldsymbol{x}_s^{l+1}=\sum_{i=-k}^{k} f_i(\boldsymbol{x}_{s+i}^{l})=\sum_{i=-k}^{k} h_i \circ g_i (\boldsymbol{x}_{s+i}^{l}),
$$
where $g_i$ is a spatial convolution to transform features $\boldsymbol{x}_{s+i}$ and $h_i$ is an operator to adjust height and width of features towards target scale $s$. The scale-wise convolution acts as a sliding window to perform the same transformation along the scale dimension. Such a convolution operation is designed to extract information from features on multiple scales in a compact manner, and is able to better capture scale-invariance than naive multi-scale or single-scale networks, which will be shown in our experiments.

To reduce additional parameters of scale-wise convolution compared with legacy spatial one, the operation $f_i$ can be further decomposed as
$$
f_i = h_i \circ g_i = h_i \circ q_i \circ p,
$$
where $p$ is a spatial convolution with shared kernels across scales and $q_i$ is a point-wise convolution for scale-specific feature transformation. Hence, the only additional parameters in $q_i$ are negligible.

There are multiple candidate operations to achieve features from different scales, including (1) convolution and deconvolution with stride, (2) average pooling and nearest neighbour up-scaling, (3) bilinear re-scaling for both down-scaling and up-scaling. In convolution, deconvolution and average pooling, the ratio of up-scaling and down-scaling is controlled by the striding, which is an integer number (\eg stride=2). In comparison, the bilinear re-scaling approach are more flexible since it can be applied to arbitrary scale ratios. We will also show in experiments that the bilinear re-scaling achieves slightly better performance. Therefore, we use bilinear re-scaling as our feature fusing operator by default.

\subsubsection{Comparison to Spatial Convolution.}
Regular spatial convolution explores context information from neighbouring pixels. In a convolutional neural network (assuming no pooling operations), the receptive field grows \textit{linearly} by the increase of the convolutional layers. Our proposed scale-wise convolution is built on a multi-scale feature pyramid. It mutually exchanges cross-scale information in every layer, and the spatial receptive field grows \textit{exponentially} because the information of smaller scales are progressively fused into the current scale.

\subsection{Scale-wise Convolutional Networks for Image Restoration}

The proposed SCN is built upon widely-activated residual networks for image super-resolution~\cite{yu2018wide,fan2018wide,fan2019empirical}, by adding multi-scale features and scale-wise convolutions in residual blocks.

\begin{figure}
  \centering
  \includegraphics[width=1.0\columnwidth]{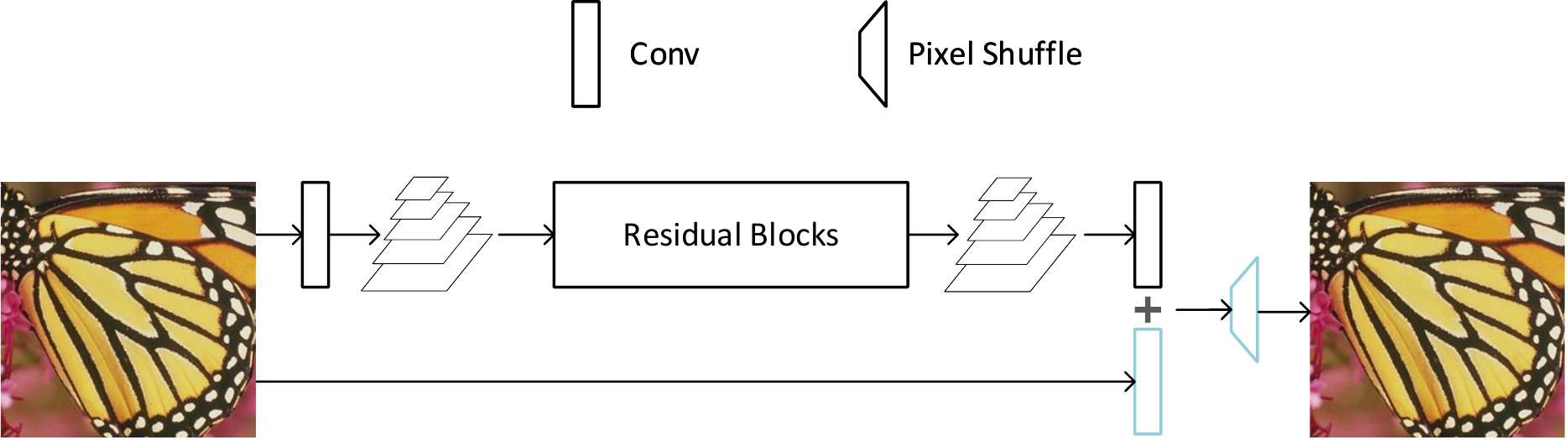}
  \caption{Overview of scale-wise convolutional networks for image restoration. Input image is first transformed into feature pyramid, then processed by multiple residual blocks with scale-wise convolution inside, finally feature pyramid is converted into target scale and fused with global skipped features.}
  \label{fig:scn}
\end{figure}


\subsubsection{Unified Architecture for Image Restoration.} 
The proposed SCN has multiple cascaded residual blocks and a global skip path way, as shown in Figure~\ref{fig:scn}.
Many low-level image restoration problems can be treated as dense pixel prediction task and share the same unified network structure. For image super-resolution, additional pixel shuffle layer and spatial convolution layer for global skip connection (blue modules in Figure~\ref{fig:scn}) are required to map low-resolution input to high-resolution counterpart. For the task of image denoising and JPEG image compression artifacts removal, the additional layers in blue are removed.

\subsubsection{Multi-Scale Feature Pyramid.}
The scale-wise convolution is performed on a consecutive multi-scale feature pyramid. In SCN, the initial feature pyramid is built from a linearly transformed features of original images (\ie mapping input image to feature space with a simple spatial convolution layer). As shown in Figure~\ref{fig:scn}, the feature maps after the first spatial convolution layer are progressively down-sampled using the re-scaling operations (bi-linear re-scaling by default). After multiple scale-wise convolutional residual blocks, each feature scale will have the information from distant scales and a exponentially increased spatial receptive field.

The scaling factor for building feature pyramid needs to be chosen carefully. When factor is close to one, features from adjacent scales will be similar and redundant. Meanwhile, receptive field growth will also degrade from exponential to linear by binomial expansion. When factor is close to zero, the number of scales will be limited.

\begin{figure*}[htb]
\centering
\includegraphics[width=2.1\columnwidth]{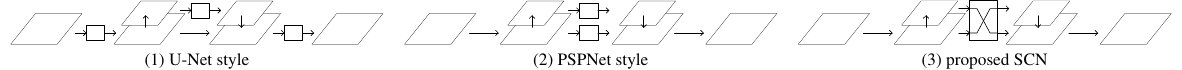}
\caption{Illustration of multi-scale network architectures.}
\label{fig:multiscale}
\end{figure*}

\subsubsection{Residual Blocks with Scale-wise Convolution.} 
Based on the feature pyramid, we apply the same function (\ie several shared layers) to process different scales. Each residual block has a spatial convolution layer for expanding features, a ReLU activation layer and another spatial convolution layer for mapping feature back to original feature size.
Afterwards we aggregate the neighbouring scales, as shown in Figure~\ref{fig:scale_conv}. The proposed scale-wise convolution learns to dynamically activate and aggregate features from different input scales in each residual building block. 

Widely-activated residual networks~\cite{yu2018wide} with wider features before ReLU activation achieve significantly better performance for image and video super-resolution, with the same parameters and computational budgets. The resulted residual network has a slim identity mapping pathway with wider channels before activation in each residual block. 
The efficiency of wide activation could further make our SCN compact. Moreover, to fully utilize the benefits of wide activation, we put up-scaling / down-scaling after the second spatial-wise convolution, where the channel numbers have been reduced. It speeds up the running time of our SCN.

\subsubsection{Comparison to Multi-Scale Architectures.} 
Our proposed SCN, at the first glance, is similar to multi-scale neural networks. However, they are fundamentally different in two aspects. 

First, the scale-wise convolution applies same weights to different scales, while other multi-scale networks usually aggregate features in difference scales with specific parameters. 
For example, U-Net style~\cite{ronneberger2015u} models in Figure~\ref{fig:multiscale} progressively down-sample features through explicit stages with multiple network layers in encoder and linearly composite multi-scale features by skip connections in decoder.
Moreover, our proposed scale-wise convolution also adapts a unified operator to fuse between different scales and thus is more compact. 

Second, the proposed scale-wise convolution aggregates multi-scale features gradually and locally, while other multi-scale networks fuse features at specific layers. 
For example, PSPNet style~\cite{zhao2017pyramid} models in Figure~\ref{fig:multiscale} independently execute over scales and combine multi-scale features in the final layers.
The scale-wise convolution fully utilizes the similarity prior of images across scales and is more robust to scale variance. Thus, the proposed operator can be viewed as ``convolution'' across scales.

Our experiments show that the proposed SCN achieved better performance than previous multi-scale architectures.


\section{Experimental Results}
\label{sec:exp}

We performance our experiments on image super-resolution, denoising and compression artifact removal to show the significance of our proposed SCN for image restoration.

\subsection{Experimental Settings}
\subsubsection{Dataset.} 
Multiple datasets are used for different image restoration tasks.

For image super-resolution, models are trained on DIV2K \cite{timofte2017ntire} dataset with 800 high resolution images since the dataset is relatively large and contains high-quality (2K resolution) images. The default splits of DIV2K dataset consist 800 training images and 100 validation images. The benchmark evaluation sets include Set5~\cite{bevilacqua2012low}, Set14~\cite{zeyde2010single}, BSD100~\cite{martin2001database} and Urban100~\cite{huang2015single} with three upscaling factors: x2, x3 and x4.

For image denoising, we use Berkeley Segmentation Dataset (BSD)~\cite{martin2001database} 200 training and 200 testing images for training purpose, as~\cite{zhang2017beyond}. The benchmark evaluation sets are Set12, BSD64~\cite{martin2001database} and Urban100~\cite{huang2015single} with noise level of 15, 25, 50.

For compression artifacts removal, we use 91 images in \cite{yang2010image} and 200 training images in \cite{martin2001database} for training. The benchmark evaluation sets are LIVE1 and Classic5 with JPEG compression quality 10, 20, 30 and 40.

\subsubsection{Training Setting.} Experiments are conducted in patch-based images and their degraded counterparts. Data augmentation including flipping and rotation are performed online during training, and Gaussian noise for image denoising tasks is also online sampled. 1000 patches randomly sample per image and per epoch, and 40 epochs in total. All models are trained with L1 distance through ADAM optimizer, and learning rate starts from 0.001 and halves every 3 epochs after the 25th. We use deep models with 8 residual blocks, 32 residual units and 4x width multiplier for most experiments, and 64 blocks and 64 units for super-resolution benchmarks.

\subsection{Main Results}
In this part, we compare our models with the state-of-the-art methods on image super-resolution, image denoising and image compression artifact removal. The results are shown in Table~\ref{tabs:sisr} (Single Image Super-Resolution), Table~\ref{table:denoising1} (Image Denoising) and Table~\ref{tab:results_psnr_ssim_car_y} (Image Compression Artifact Removal). 

\subsubsection{Image super-resolution.}
We compare our SCN with state-of-the-art single image super-resolution methods: A+~\cite{timofte2014a+}, SRCNN~\cite{dong2014learning}, VDSR~\cite{kim2016accurate}, EDSR~\cite{lim2017enhanced}, WDSR~\cite{yu2018wide}. Self-ensemble strategy is also used to further improve our SCN, which we denote as SCN+. 

As shown in Table~\ref{tabs:sisr}, our proposed SCN and SCN+ achieves best results on all benchmark dataset and across all three upscaling factors. It indicates that properly modeling scale-invariance into neural networks, specifically the scale-wise convolution, can bring significant benefits to image super-resolution.

Furthermore, we visually compare our super-resolved images with other baseline methods. As shown in the Figure~\ref{fig:sr_vis}, our method produces higher quality images (for example, sharper edges) than our baseline methods: SRCNN~\cite{dong2014learning}, FSRCNN~\cite{dong2016accelerating}, VDSR~\cite{kim2016accurate}, LapSRN~\cite{lai2017deep}, MemNet~\cite{tai2017memnet} and EDSR~\cite{lim2017enhanced}.

\begin{table*}[!htbp]
        \caption{Public image super-resolution benchmark results and DIV2K validation results in PSNR / SSIM.
			Red indicates the best performance and blue indicates the second best. \(+\) indicates results with self-ensemble.}
		\label{tabs:sisr}
		{
			\renewcommand{\arraystretch}{1.4}
			\setlength\tabcolsep{5pt}
			\begin{center}
			\begin{adjustbox}{width=2.1\columnwidth}
					\begin{tabular} {|*{10}{c|}}
						\hline
						&&&&& &&&& \\ [-1em]
						Dataset & Scale & Bicubic &
						A+ &
						SRCNN &
						VDSR &
						\parbox[c]{1.cm}{\centering EDSR}& 
						\parbox[c]{1.cm}{\centering WDSR} & 
						\parbox[c]{1.cm}{\centering\textbf{SCN }} & 
						\parbox[c]{1.cm}{\centering\textbf{SCN+ }} \\ 
						&&&&& &&&&\\ [-1em]
						\hline
						\hline
						& $\times 2$ &
						33.66 / 0.9299 & 36.54 / 0.9544 & 36.66 / 0.9542 & 37.53 / 0.9587 &
						37.99 / 0.9604 & 
						38.10 / 0.9608 &
						\textcolor{blue}{38.18} / \textcolor{blue}{0.9614} & 
						\textcolor{red}{38.29} / \textcolor{red}{0.9616} \\
						Set5 & $\times 3$ &
						30.39 / 0.8682 & 32.58 / 0.9088 & 32.75 / 0.9090 & 33.66 / 0.9213 &
						34.37 / 0.9270 & 
						34.48 / 0.9279 &
						\textcolor{blue}{34.60} / \textcolor{blue}{0.9295} &
						\textcolor{red}{34.75} / \textcolor{red}{0.9301} \\
						& $\times 4$ &
						28.42 / 0.8104 & 30.28 / 0.8603 & 30.48 / 0.8628 & 31.35 / 0.8838 & 
						32.09 / 0.8938 & 
						32.27 / 0.8963 &
						\textcolor{blue}{32.39} / \textcolor{blue}{0.8981} & 
						\textcolor{red}{32.59} / \textcolor{red}{0.9000} \\
						\hline
						& $\times 2$ &
						30.24 / 0.8688 & 32.28 / 0.9056 & 32.42 / 0.9063 & 33.03 / 0.9124 &
						33.57 / 0.9175 & 
						33.72 / 0.9182 &
						\textcolor{blue}{33.99} / \textcolor{blue}{0.9208} & 
						\textcolor{red}{34.14} / \textcolor{red}{0.9218} \\
						Set14 & $\times 3$ &
						27.55 / 0.7742 & 29.13 / 0.8188 & 29.28 / 0.8209 & 29.77 / 0.8314 &
						30.28 / 0.8418 & 
						30.39 / 0.8434 &
						\textcolor{blue}{30.50} / \textcolor{blue}{0.8467} & 
						\textcolor{red}{30.62} / \textcolor{red}{0.8483} \\
						& $\times 4$ &
						26.00 / 0.7027 & 27.32 / 0.7491 & 27.49 / 0.7503 & 28.01 / 0.7674 &
						28.58 / 0.7813 & 
						28.67 / 0.7838 &
						\textcolor{blue}{28.74} / \textcolor{blue}{0.7869} & 
						\textcolor{red}{28.90} / \textcolor{red}{0.7895} \\
						\hline
						& $\times 2$ &
						29.56 / 0.8431 & 31.21 / 0.8863 & 31.36 / 0.8879 & 31.90 / 0.8960 &
						32.16 / 0.8994 & 
						32.25 / 0.9004 &
						\textcolor{blue}{32.39} / \textcolor{blue}{0.9024} & 
						\textcolor{red}{32.43} / \textcolor{red}{0.9028} \\						
						B100 & $\times 3$ &
						27.21 / 0.7385 & 28.29 / 0.7835 & 28.41 / 0.7863 & 28.82 / 0.7976 &
						29.09 / 0.8052 & 
						29.16 / 0.8067 &
						\textcolor{blue}{29.26} / \textcolor{blue}{0.8104} & 
						\textcolor{red}{29.34} / \textcolor{red}{0.8115} \\
						& $\times 4$ &
						25.96 / 0.6675 & 26.82 / 0.7087 & 26.90 / 0.7101 & 27.29 / 0.7251 & 
						27.57 / 0.7357 & 
						27.64 / 0.7383 &
						\textcolor{blue}{27.69} / \textcolor{blue}{0.7415} & 
						\textcolor{red}{27.79} / \textcolor{red}{0.7436} \\
						\hline
						& $\times 2$ &
						26.88 / 0.8403 & 29.20 / 0.8938 & 29.50 / 0.8946 & 30.76 / 0.9140 &
						31.98 / 0.9272 & 
						32.37 / 0.9302 &
						\textcolor{blue}{33.13} / \textcolor{blue}{0.9374} & 
						\textcolor{red}{33.26} / \textcolor{red}{0.9386} \\
						Urban100 & $\times 3$ &
						24.46 / 0.7349 & 26.03 / 0.7973 & 26.24 / 0.7989 & 27.14 / 0.8279 &
						28.15 / 0.8527 & 
						28.38 / 0.8567 &
						\textcolor{blue}{28.79} / \textcolor{blue}{0.8667} & 
						\textcolor{red}{29.01} / \textcolor{red}{0.8700} \\
						& $\times 4$ &
						23.14 / 0.6577 & 24.32 / 0.7183 & 24.52 / 0.7221 & 25.18 / 0.7524 & 
						26.04 / 0.7849 & 
						26.26 / 0.7911 &
						\textcolor{blue}{26.50} / \textcolor{blue}{0.8000} & 
						\textcolor{red}{26.76} / \textcolor{red}{0.8055} \\
						\hline\hline
						\multirow{3}{*}{\parbox[c]{2cm}{\centering DIV2K\\validation}}
						& $\times 2$ &
						31.01 / 0.8923 & 32.89 / 0.9180 & 33.05 / - & 33.66 /  0.9290 &
						34.61 / 0.9372 &	
						34.78 / 0.9384 &
						\textcolor{blue}{35.10} / \textcolor{blue}{0.9411} &
						\textcolor{red}{35.19} / \textcolor{red}{0.9413} \\
						& $\times 3$ &
						28.22 / 0.8124  & 29.50 / 0.8440 & 29.64 / - & 30.09 / 0.8590 &
						30.92 / 0.8734 &	
						31.04 / 0.8755 &
						\textcolor{blue}{31.28} / \textcolor{blue}{0.8800} &
						\textcolor{red}{31.39} / \textcolor{red}{0.8814} \\
						& $\times 4$ &
						26.66 / 0.7512 & 27.70 / 0.7840 & 27.78 / - & 28.17 / 0.8000 &
						28.95 / 0.8178 &
						29.06 / 0.8213 &
						\textcolor{blue}{29.18} / \textcolor{blue}{0.8253} &
						\textcolor{red}{29.36} / \textcolor{red}{0.8286} \\
						\hline								
					\end{tabular}
				\end{adjustbox}
			\end{center}
		}
		
	\end{table*}

\subsubsection{Image denoising.}
We compare our SCN with state-of-the-art image denoising methods: BM3D~\cite{dabov2007video}, WNNM~\cite{gu2014weighted}, TNRD~\cite{chen2016trainable} and DnCNN~\cite{zhang2017beyond}. Table~\ref{table:denoising1} shows that our proposed SCN achieves higher quantitative results in both PSNR and SSIM, on all benchmark dataset and across all three noise levels. 

\begin{table}[tb]
	\centering
		\caption{Benchmark image denoising results. Training and testing protocols are followed as in~\cite{zhang2017beyond}. Average PSNR/SSIM for various noise levels. The best results are in bold. }
		\label{table:denoising1}		
		\resizebox{1.0\columnwidth}{!}{
		\begin{tabular}{|*{7}{c|}}
			\hline
			Dataset & Noise & BM3D & WNNM & TNRD & DnCNN & SCN \\ 
			\hline 
			\hline
			\multirow{3}{*}{Set12} 
			& 15 & 32.37/0.8952 & 32.70/0.8982 & 32.50/0.8958 & 32.86/0.9031 & \textbf{32.99}/\textbf{0.9055} \\ 
			& 25 & 39.97/0.8504 & 30.28/0.8557 & 30.06/0.8512 & 30.44/0.8622 & \textbf{30.64}/\textbf{0.8677} \\ 
            & 50 & 26.72/0.7676 & 27.05/0.7775 & 26.81/0.7680 & 27.18/0.7829 & \textbf{27.43}/\textbf{0.7967} \\ 
			\hline
            \hline
			\multirow{3}{*}{BSD68}
			& 15 & 31.07/0.8717 & 31.37/0.8766 & 31.42/0.8769 & 31.73/0.8907 & \textbf{31.80}/\textbf{0.8933} \\ 
			& 25 & 28.57/0.8013 & 28.83/0.8087 & 28.92/0.8093 & 29.23/0.8278 & \textbf{29.31}/\textbf{0.8329} \\ 
            & 50 & 25.62/0.6864 & 25.87/0.6982 & 25.97/0.6994 & 26.23/0.7189 & \textbf{26.34}/\textbf{0.7296} \\
			\hline
            \hline
			\multirow{3}{*}{Urban100} 
			& 15 & 32.35/0.9220 & 32.97/0.9271 & 31.86/0.9031 & 32.68/0.9255 & \textbf{32.99}/\textbf{0.9304} \\ 
            & 25 & 29.70/0.8777 & 30.39/0.8885 & 29.25/0.8473 & 29.97/0.8797 & \textbf{30.39}/\textbf{0.8911} \\ 
			& 50 & 25.95/0.7791 & 26.83/0.8047 & 25.88/0.7563 & 26.28/0.7874 & \textbf{26.84}/\textbf{0.8150} \\ 
			\hline		
		\end{tabular}
		}
\end{table}

\subsubsection{Image compression artifact removal.}
Further, we compare SCN with state-of-the-art image compression artifact removal methods: JPEG, SA-DCT~\cite{foi2007pointwise}, ARCNN~\cite{dong2015compression}, TNRD~\cite{chen2016trainable} and DnCNN~\cite{zhang2017beyond}. Our SCN constantly outperforms the baseline methods in Table~\ref{tab:results_psnr_ssim_car_y}, on all benchmark dataset and across different JPEG compression qualities.

\begin{table}[tb]
\caption{Compression artifacts reduction benchmark results. Best results are in bold.}
\label{tab:results_psnr_ssim_car_y}

\resizebox{1.0\columnwidth}{!}{
\begin{tabular}{|*{8}{c|}}
\hline
Dataset & $q$ &  JPEG  & SA-DCT &  ARCNN &  TNRD &  DnCNN &  SCN
\\
\hline
\hline
\multirow{4}{*}{LIVE1} & 10 
& 27.77 / 0.7905
  & 28.65 / 0.8093
    & 28.98 / 0.8217
      & 29.15 / 0.8111
        & {29.19} / {0.8123}
          & \textbf{29.36} / \textbf{0.8179}
\\
& 20 
& 30.07 / 0.8683
  & 30.81 / 0.8781
    & 31.29 / 0.8871
      & 31.46 / 0.8769
        & {31.59} / {0.8802}
          & \textbf{31.66} / \textbf{0.8825}
\\
& 30 
& 31.41 / 0.9000
  & 32.08 / 0.9078
    & 32.69 / 0.9166
      & 32.84 / 0.9059
        & {32.98} / {0.9090}
          & \textbf{33.04} / \textbf{0.9106}
\\
& 40 
& 32.35 / 0.9173
  & 32.99 /  0.9240
    & 33.63 / 0.9306
      & - / -
        & {33.96} / {0.9247}
          & \textbf{34.02} / \textbf{0.9263}
\\
\hline 
\hline
\multirow{4}{*}{Classic5} & 10 
& 27.82 / 0.7800
  & 28.88 / 0.8071
    & 29.04 / 0.8111
      & 29.28 / 0.7992
        & {29.40} / {0.8026}
          & \textbf{29.53} / \textbf{0.8081}
\\
& 20 
& 30.12 / 0.8541
  & 30.92 / 0.8663
    & 31.16 / 0.8694
      & 31.47 / 0.8576
        & {31.63} / {0.8610}
          & \textbf{31.72} / \textbf{0.8637}
\\
& 30 
& 31.48 / 0.8844
  & 32.14 / 0.8914
    & 32.52 / 0.8967
      & 32.78 / 0.8837
        & {32.91} / {0.8861}
          & \textbf{32.92} / \textbf{0.8873}
\\
& 40 
& 32.43 / 0.9011
  & 33.00 / 0.9055
    & 33.34 / 0.9101
      & - / -
        & {33.77} / {0.9003}
          & \textbf{33.78} / \textbf{0.9022}
\\
\hline          
\end{tabular}

}
\end{table}

\newcommand{\widthscalefive}{0.145}
\newlength\fsdurthree
\setlength{\fsdurthree}{0.2mm}
\begin{figure*}
    \centering
    \begin{minipage}{1.0\textwidth}
\begin{adjustbox}{valign=t}
		\tiny
			\begin{tabular}{@{}c}	\includegraphics[width=0.229\textwidth, trim={525 100 0 200},clip]{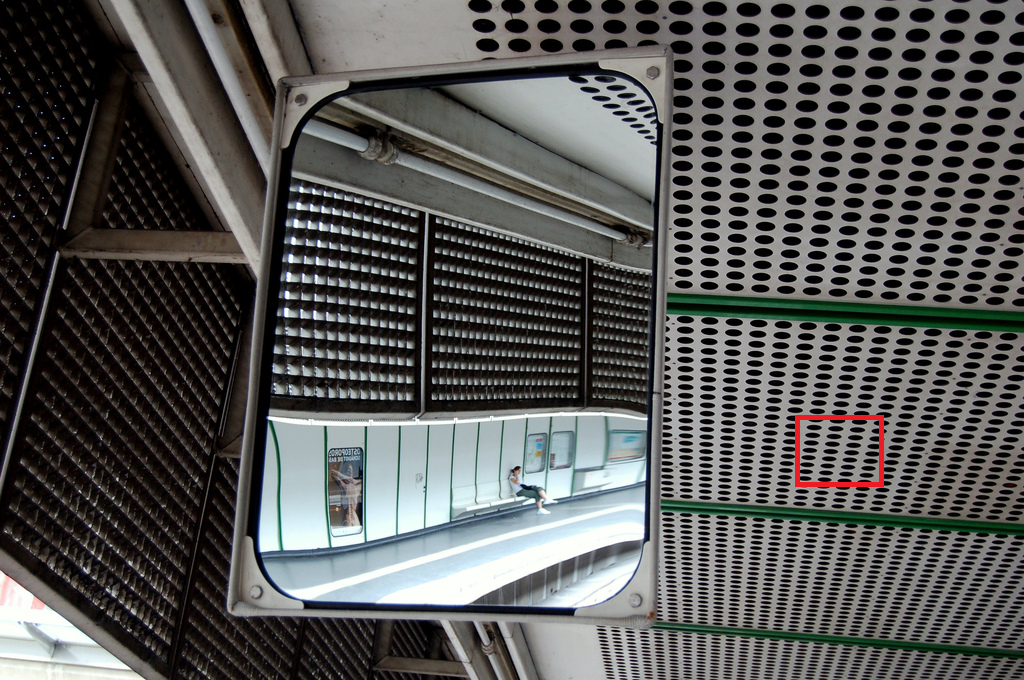}
				\\
				Urban100 ($\times 4$):
				\\
				img\_004
			\end{tabular}
		\end{adjustbox}
		\hspace{-2.3mm}
        \begin{adjustbox}{valign=t}
		\tiny
			\begin{tabular}{@{}c@{}c@{}c@{}c@{}c@{}c@{}}
				\includegraphics[width=\widthscalefive \textwidth]{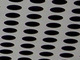} \hspace{\fsdurthree} &
				\includegraphics[width=\widthscalefive \textwidth]{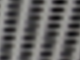} \hspace{\fsdurthree} &
				\includegraphics[width=\widthscalefive \textwidth]{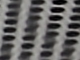} \hspace{\fsdurthree} &
				\includegraphics[width=\widthscalefive \textwidth]{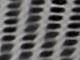} \hspace{\fsdurthree} &
				\includegraphics[width=\widthscalefive \textwidth]{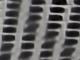} 
				\\
				HR \hspace{\fsdurthree} &
				Bicubic \hspace{\fsdurthree} &
				SRCNN \hspace{\fsdurthree} &
				FSRCNN \hspace{\fsdurthree} &
				VDSR
				\\
				\includegraphics[width=\widthscalefive \textwidth]{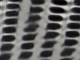} \hspace{\fsdurthree} &
				\includegraphics[width=\widthscalefive \textwidth]{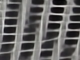} \hspace{\fsdurthree} &
				\includegraphics[width=\widthscalefive \textwidth]{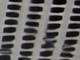} \hspace{\fsdurthree} &
				\includegraphics[width=\widthscalefive \textwidth]{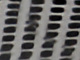} \hspace{\fsdurthree} &
				\includegraphics[width=\widthscalefive \textwidth]{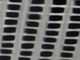}  
				\\ 
				LapSRN \hspace{\fsdurthree} &
				MemNet \hspace{\fsdurthree} &
				EDSR \hspace{\fsdurthree} &
				SRMDNF \hspace{\fsdurthree} &
				SCN 
				\\
			\end{tabular}
\end{adjustbox}
\end{minipage} \\
    \begin{minipage}{1.0\textwidth}
\begin{adjustbox}{valign=t}
		\tiny
			\begin{tabular}{@{}c}	\includegraphics[width=0.229\textwidth, trim={0 300 500 0},clip]{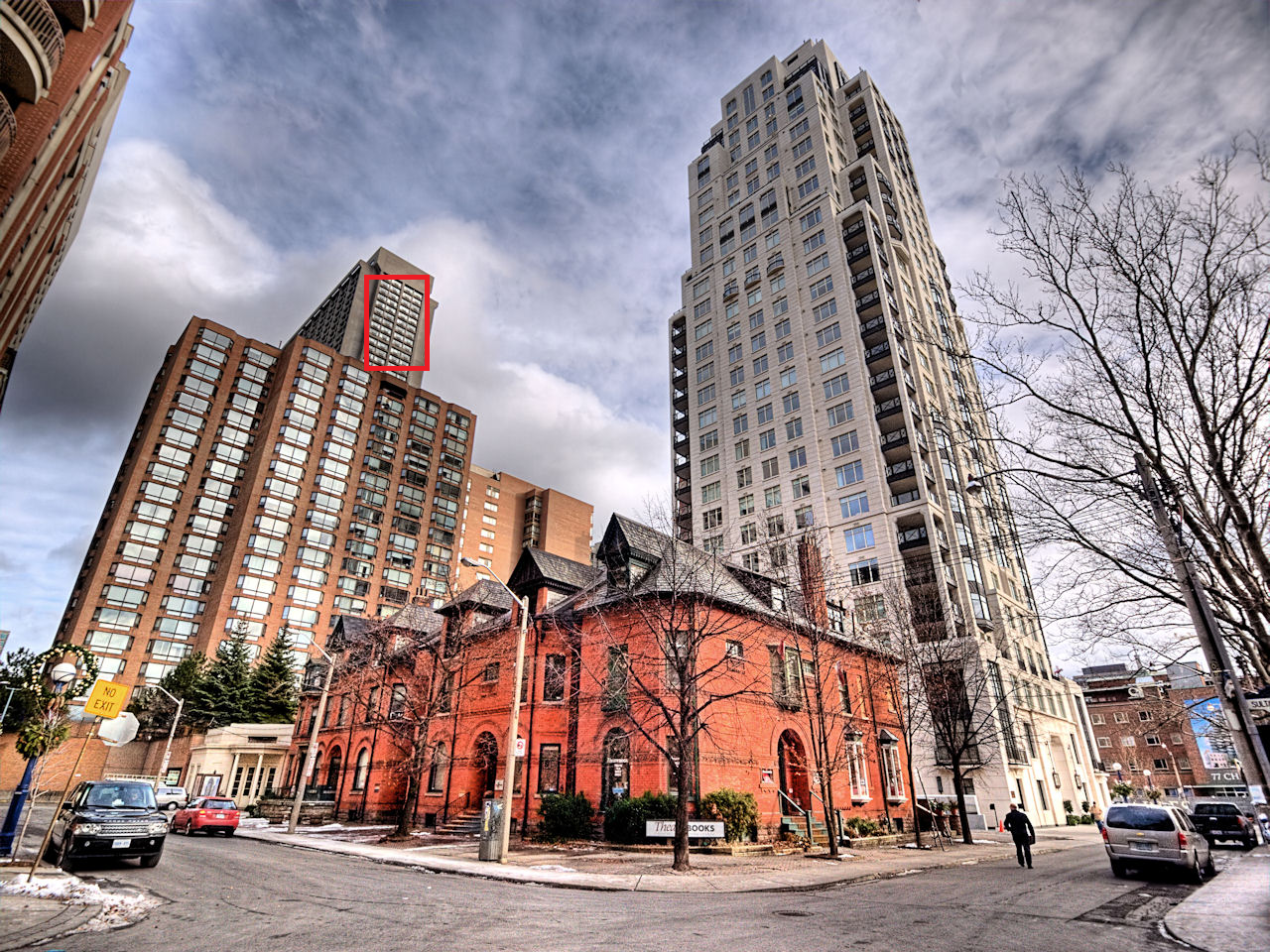}
				\\
				Urban100 ($\times 4$):
				\\
				img\_020
			\end{tabular}
		\end{adjustbox}
		\hspace{-2.3mm}
        \begin{adjustbox}{valign=t}
		\tiny
			\begin{tabular}{@{}c@{}c@{}c@{}c@{}c@{}c@{}}
				\includegraphics[width=\widthscalefive \textwidth]{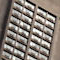} \hspace{\fsdurthree} &
				\includegraphics[width=\widthscalefive \textwidth]{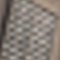} \hspace{\fsdurthree} &
				\includegraphics[width=\widthscalefive \textwidth]{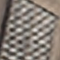} \hspace{\fsdurthree} &
				\includegraphics[width=\widthscalefive \textwidth]{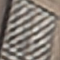} \hspace{\fsdurthree} &
				\includegraphics[width=\widthscalefive \textwidth]{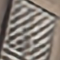} 
				\\
				HR \hspace{\fsdurthree} &
				Bicubic \hspace{\fsdurthree} &
				SRCNN \hspace{\fsdurthree} &
				FSRCNN \hspace{\fsdurthree} &
				VDSR 
				\\
				\includegraphics[width=\widthscalefive \textwidth]{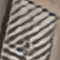} \hspace{\fsdurthree} &
				\includegraphics[width=\widthscalefive \textwidth]{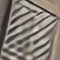} \hspace{\fsdurthree} &
				\includegraphics[width=\widthscalefive \textwidth]{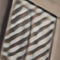} \hspace{\fsdurthree} &
				\includegraphics[width=\widthscalefive \textwidth]{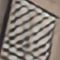} \hspace{\fsdurthree} &
				\includegraphics[width=\widthscalefive \textwidth]{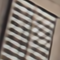}  
				\\ 
				LapSRN \hspace{\fsdurthree} &
				MemNet \hspace{\fsdurthree} &
				EDSR \hspace{\fsdurthree} &
				SRMDNF \hspace{\fsdurthree} &
				SCN 
				\\
			\end{tabular}
\end{adjustbox}
\end{minipage} \\
    \begin{minipage}{1.0\textwidth}
\begin{adjustbox}{valign=t}
		\tiny
			\begin{tabular}{@{}c}	\includegraphics[width=0.229\textwidth, trim={0 0 300 100},clip]{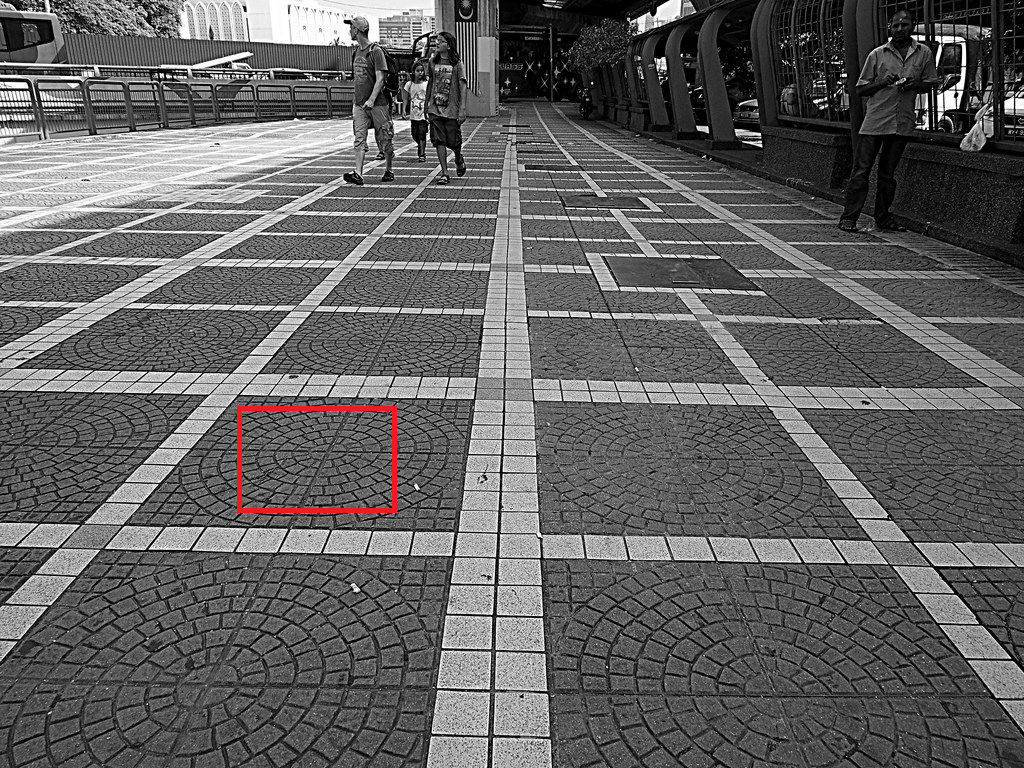}
				\\
				Urban100 ($\times 4$):
				\\
				img\_095
			\end{tabular}
		\end{adjustbox}
		\hspace{-2.3mm}
        \begin{adjustbox}{valign=t}
		\tiny
			\begin{tabular}{@{}c@{}c@{}c@{}c@{}c@{}c@{}}
				\includegraphics[width=\widthscalefive \textwidth]{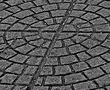} \hspace{\fsdurthree} &
				\includegraphics[width=\widthscalefive \textwidth]{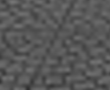} \hspace{\fsdurthree} &
				\includegraphics[width=\widthscalefive \textwidth]{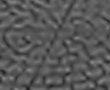} \hspace{\fsdurthree} &
				\includegraphics[width=\widthscalefive \textwidth]{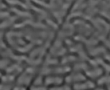} \hspace{\fsdurthree} &
				\includegraphics[width=\widthscalefive \textwidth]{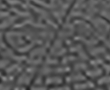} 
				\\
				HR \hspace{\fsdurthree} &
				Bicubic \hspace{\fsdurthree} &
				SRCNN \hspace{\fsdurthree} &
				FSRCNN \hspace{\fsdurthree} &
				VDSR 
				\\
				\includegraphics[width=\widthscalefive \textwidth]{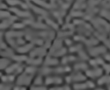} \hspace{\fsdurthree} &
				\includegraphics[width=\widthscalefive \textwidth]{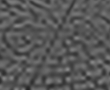} \hspace{\fsdurthree} &
				\includegraphics[width=\widthscalefive \textwidth]{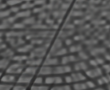} \hspace{\fsdurthree} &
				\includegraphics[width=\widthscalefive \textwidth]{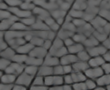} \hspace{\fsdurthree} &
				\includegraphics[width=\widthscalefive \textwidth]{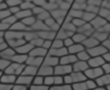}  
				\\ 
				LapSRN \hspace{\fsdurthree} &
				MemNet \hspace{\fsdurthree} &
				EDSR \hspace{\fsdurthree} &
				SRMDNF \hspace{\fsdurthree} &
				SCN 
				\\
			\end{tabular}
\end{adjustbox}
\end{minipage}
    \caption{Visual comparison results of $\times 4$ image super-resolution on Urban100 datasets. \textit{SCN} achieved qualitative results. More visual comparisons are shown in the supplementary materials.}
    \label{fig:sr_vis}
\end{figure*}

\begin{figure*}
      \centering
      \includegraphics[width=0.20\linewidth]{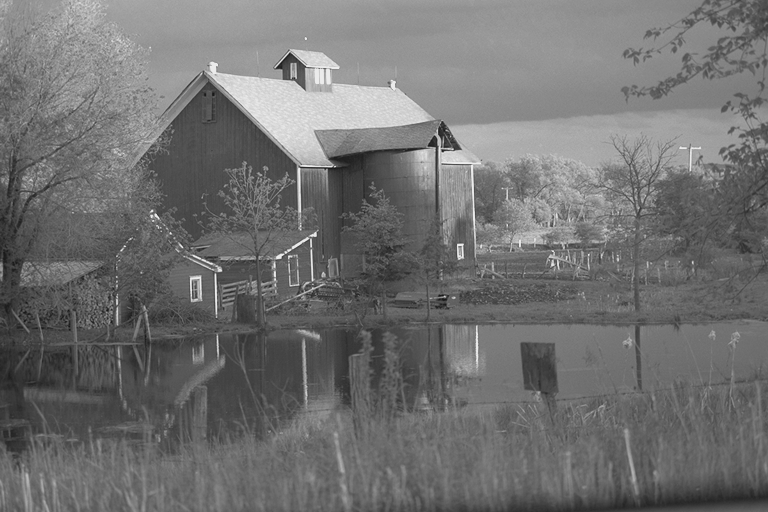}
      \includegraphics[width=0.20\linewidth]{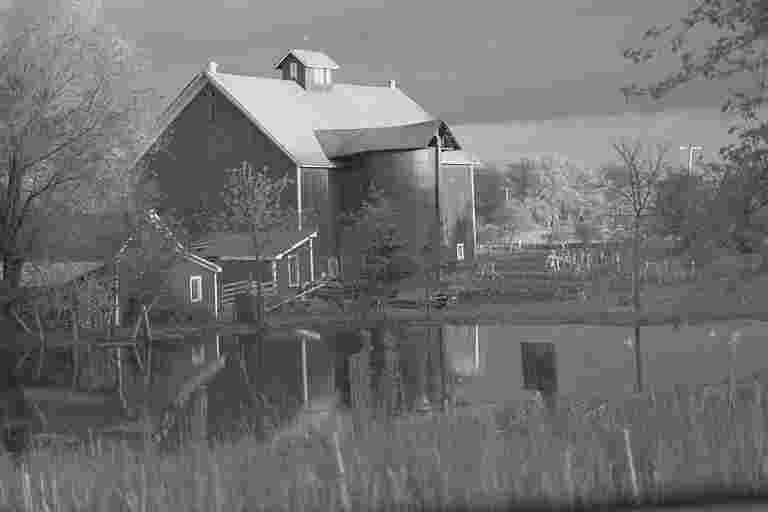}
      \includegraphics[width=0.20\linewidth]{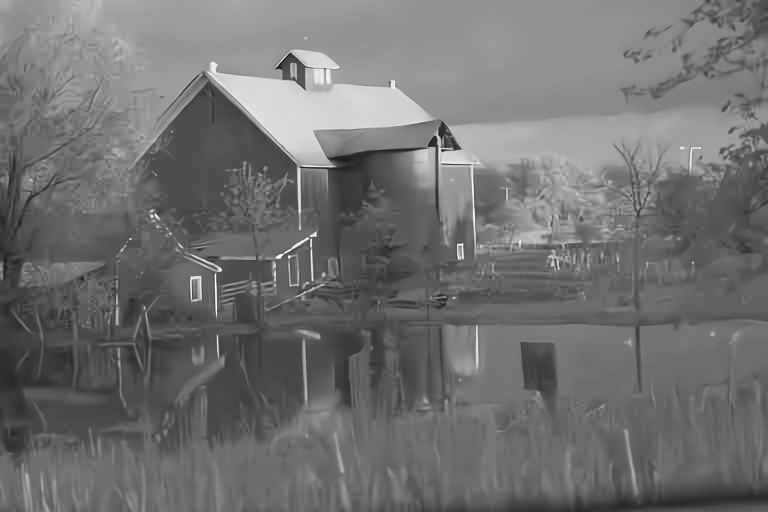} \\
      \includegraphics[width=0.20\linewidth]{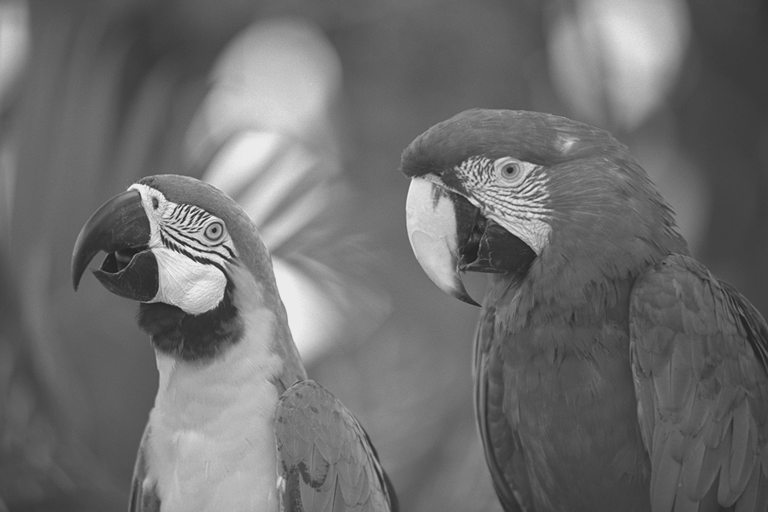}
      \includegraphics[width=0.20\linewidth]{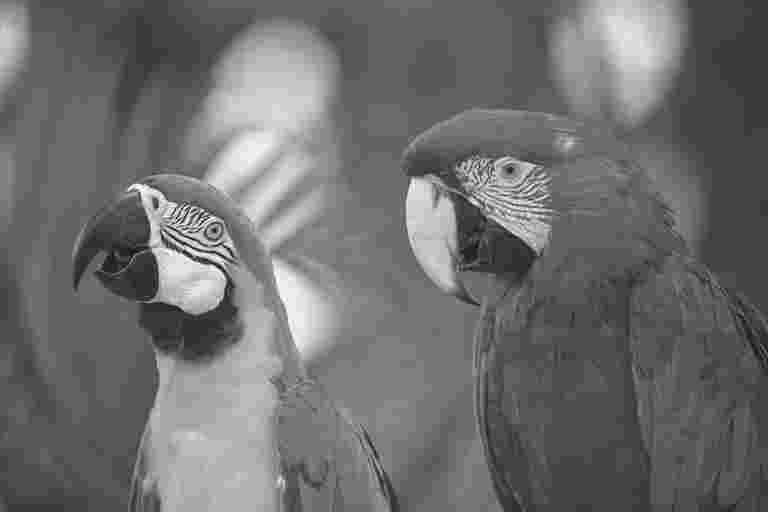}
      \includegraphics[width=0.20\linewidth]{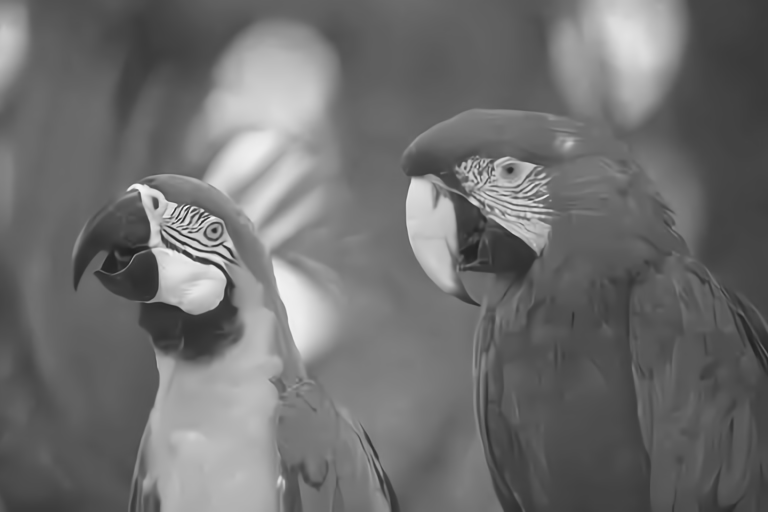}
      \caption{SCN for JPEG compression artifact removal. From left to right, we show groud truth image, compressed image, and our restored image with scale-wise convolutional network.}
      \label{fig:jpeg}
\end{figure*}

\subsection{Ablation Study}
In this section, we conduct ablation study to justify the significance of proposed scale-wise convolution. The experiments are on image super-resolution with x2 up-scaling.

\subsubsection{Multi-scale architectures.}
We compare three representative design choices of multi-scale network structures discussed in previous section.
The U-Net style model~\cite{ronneberger2015u} has 2 scales in both encoder and decoder, and 2 residual blocks per scale, then 8 blocks in total.
The PSPNet style model~\cite{zhao2017pyramid} has 2 scales and 8 residual blocks per scale, and features from scales are fused in output layer.
Our SCN also has 2 scales and 8 residual blocks.
The total numbers of parameters of the models are almost the same.
Table~\ref{tabs:multiscale} shows the advances of our proposed SCN over other multi-scale approaches.


\subsubsection{Parameter sharing across scales.}
We study whether the parameters for different scales can be shared or not. In table~\ref{tabs:sharing} we compare (1) single-scale baseline (Baseline), (2) multi-scale architecture without sharing weights (Un-shared), (3) multi-scale architecture with sharing weights (Shared) and (4) a larger model (more parameters) multi-scale architecture without sharing weights. We report their number of parameters in the Table.

Table~\ref{tabs:sharing} shows that under same number of parameters, shared version has much better performance. To achieve a similar performance, the un-shared version has to involve more parameters.
The results proof the scale-invariance of our proposed SCN and its effectiveness.

\begin{table}[tb]
    \centering
    \label{tabs:multiscale}
    \caption{Ablation study on multi-scale network structures.}
    \scriptsize
    \begin{tabular}{ |c|c| } 
     \hline
     Method & PSNR \\
     \hline
     U-Net style & 34.56 \\ 
     PSPNet style & 34.58 \\
     SCN & \textbf{34.67} \\
     \hline
    \end{tabular}

    \centering
    \caption{Ablation study on parameter sharing across scales.}
    \scriptsize
    \begin{tabular}{ |c|c|c| } 
     \hline
     Method & \# Params & PSNR \\
     \hline
     Baseline & 1.2M & 34.74 \\ 
     Un-shared  & 1.2M & 34.68 \\
     Shared  & 1.2M & \textbf{34.77} \\
     Un-shared Large & 2.1M & \textbf{34.77} \\
     \hline
    \end{tabular}
    \label{tabs:sharing}

    \centering
    \label{tabs:resampling}
    \caption{Ablation study on re-sampling methods.}
    \scriptsize
    \begin{tabular}{|c|c|c|} 
     \hline
     Down-sample & Up-sample & PSNR \\
     \hline
     Conv & Deconv & 34.77 \\ 
     AvgPool & Nearest & 34.78 \\
     Bilinear & Bilinear & \textbf{34.80} \\ 
     \hline
    \end{tabular}
\end{table}

\begin{figure}[tb]
  \centering
  \includegraphics[width=0.8\columnwidth]{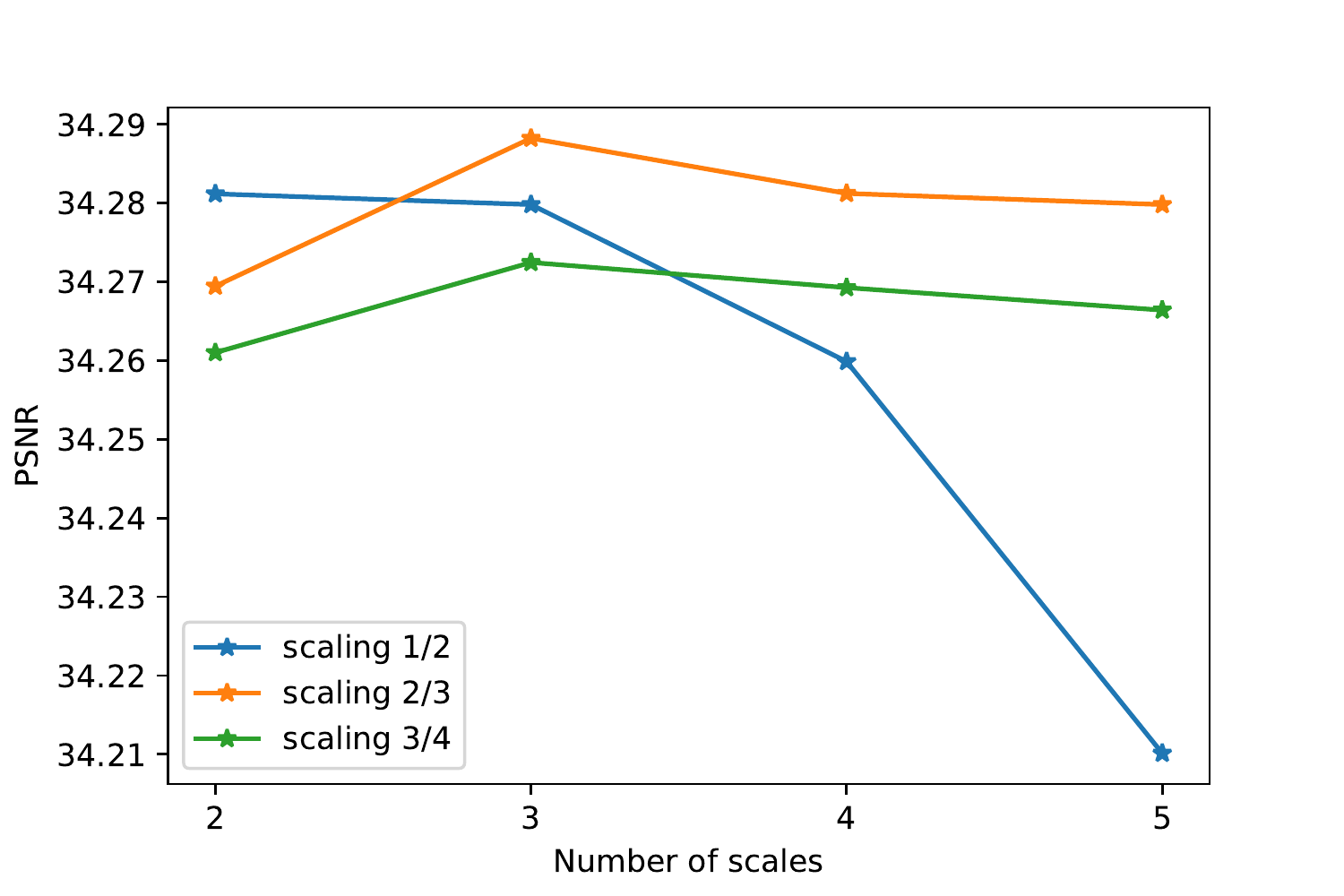}
  \caption{Ablation study on number of scales and scaling ratios.}
  \label{fig:scales}
  \vspace{-0.5cm}
\end{figure}

\begin{figure}[tb]
  \centering
  \includegraphics[width=0.8\columnwidth]{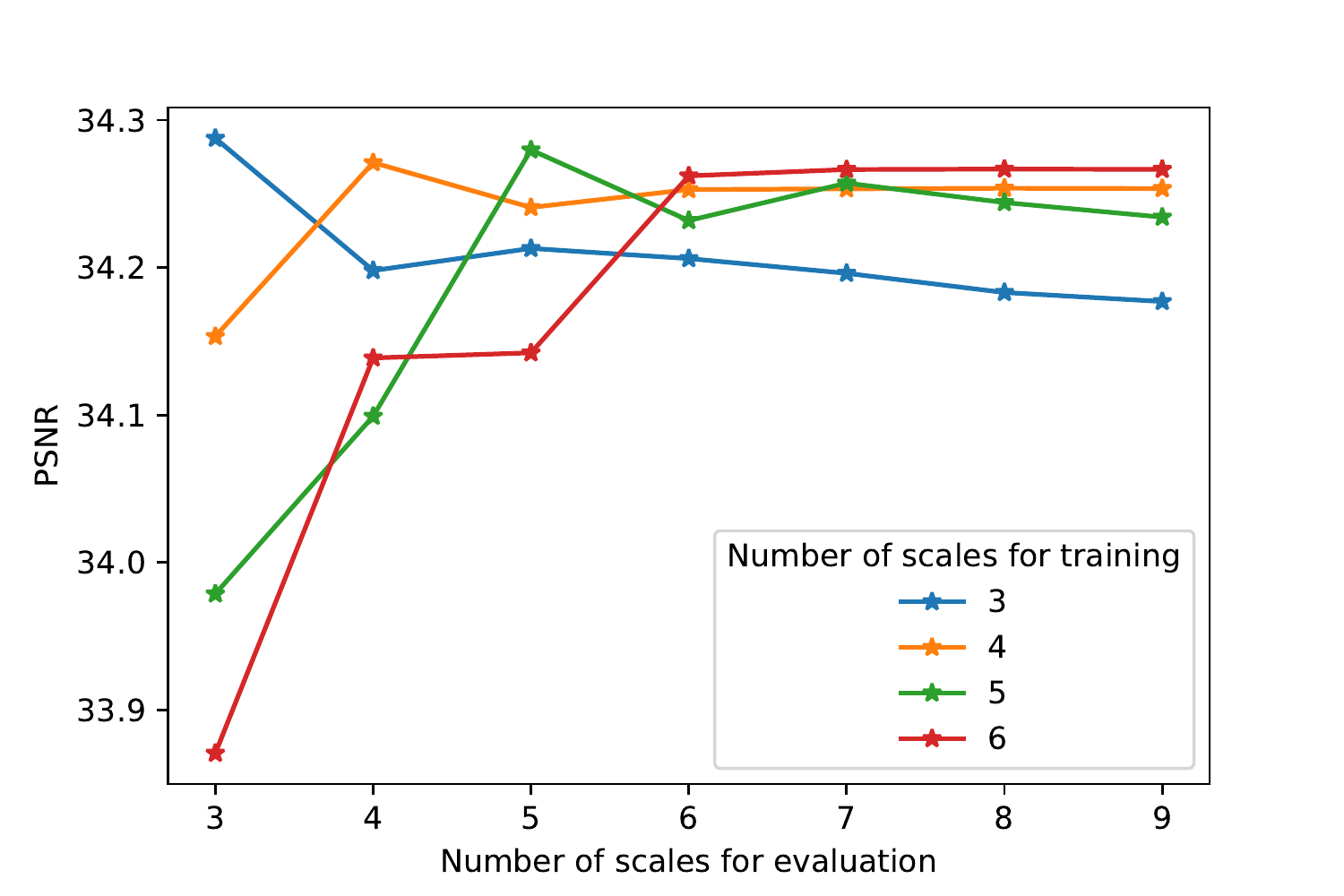}
  \caption{Ablation study on different number of scales for evaluation.}
  \label{fig:scales_n_val}
  \vspace{-0.2cm}
\end{figure}

\begin{figure}[tb]
  \centering
  \includegraphics[width=0.8\columnwidth]{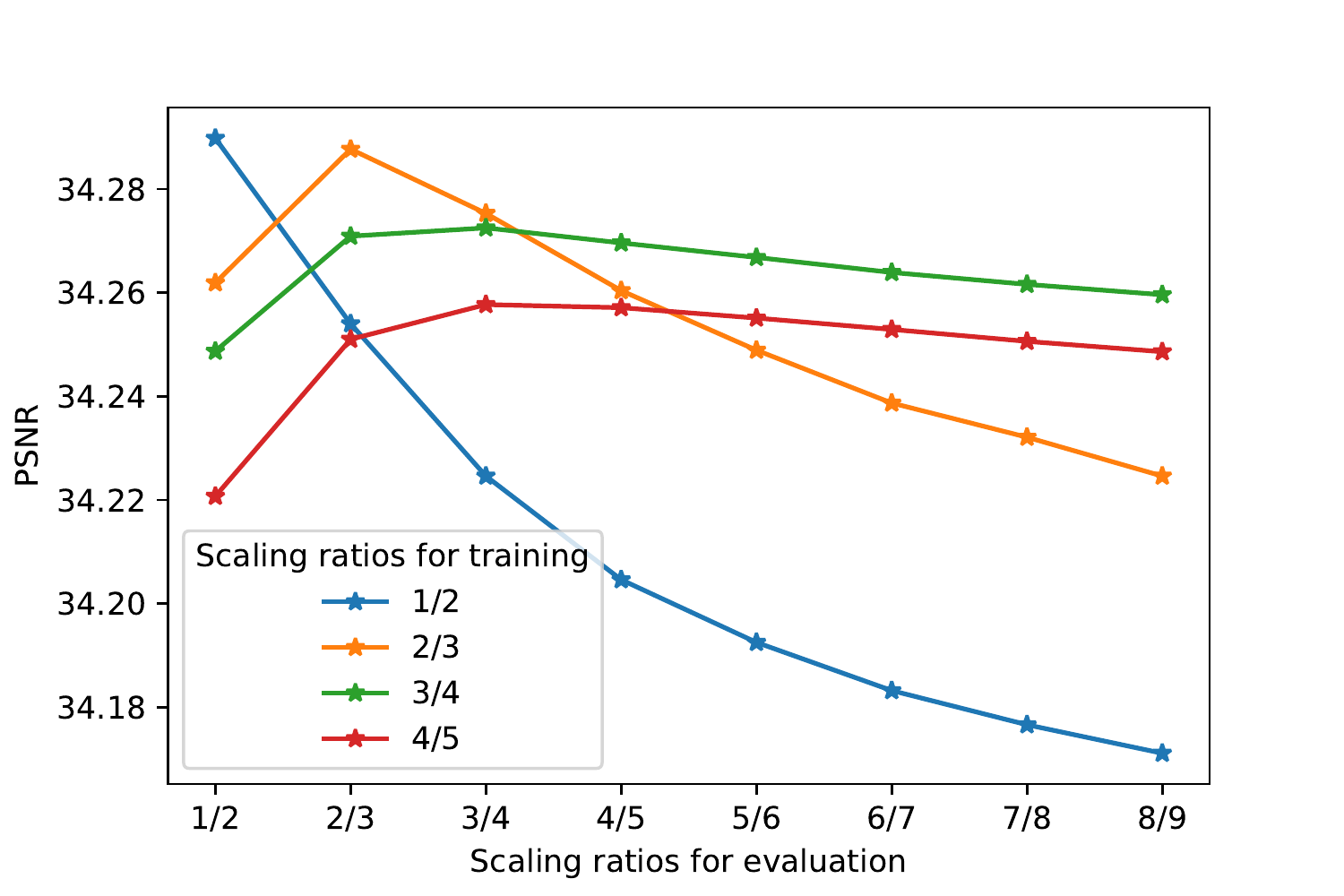}
  \caption{Ablation study on different scaling ratios for evaluation.}
  \label{fig:scales_r_val}
  \vspace{-0.5cm}
\end{figure}

\subsubsection{Resampling in Scale-wise Convolution.}
Scale-wise convolution aggregates information from nearby input scales. A resampling operation is required to fuse features of different scales into the same scale, as shown in Figure~\ref{fig:scale_conv}. In this part, we study the resampling method and compare their performances.

For resampling, there exists several methods: (1) down-sampling with strided convolution, up-sampling with strided deconvolution, (2) down-sampling with average pooling, up-sampling with nearest neighbor sampling and (3) down-sampling with bilinear sampling, up-sampling with bilinear sampling. We note that for the first two methods, the scaling ratio is limited to 2, 4, etc, because the kernel size have to be integer numbers. Our results in Table~\ref{tabs:resampling} indicate that biliear has superior performance.

\subsubsection{Scales in Scale-wise Convolution.}
In this part, we study the number of scales and the scaling ratios between neighborhood scales. From the results in Figure~\ref{fig:scales}, we have several observations. (1) The scaling ratio plays an important role in the performance: a small scaling ratio (\eg \(\frac{1}{2}\)) may lead to worse performance as the input feature map being too small, a large scaling ratio (\eg comparing the curve of \(\frac{2}{3}\) and \(\frac{3}{4}\)) impedes the growth of receptive fields. (2) Multi-scale modeling is essential for image restoration and leads to much better performance compared with single-scale (34.74 in PSNR). Meanwhile more scales may lead to slightly inferior performance, which is likely due to the noise from the smallest scale. For example, in the setting of 6 scales with \(\frac{1}{2}\) scaling ratio, the smallest image has resolution \(\frac{1}{64}\) of the original resolution.

\subsubsection{Evaluation for Scale-wise Convolution.}
The convolution on scale axis enables the SCN to be evaluated on number of scales that are different with training. We show in Figure \ref{fig:scales_n_val} that: (1) When training with limited number of scales, the performance degrades when number of scales between training and evaluation mismatch (\eg the blue curve for training with 3 scales). (2) When training with enough number of scales, the performance can be boosted even with more scales during evaluation (\eg the red curve for models trained with 6 scales, the performance grows until 9 scales). 
We also evaluated models with different scale ratios from training, results in Figure \ref{fig:scales_r_val} show that scale ratios must be matched between training and evaluation.


\section{Conclusions}
In this work we have presented scale-wise convolution that models scale-invariance in deep neural networks. The scale-wise convolutional network significantly improves the predictive accuracy on several image restoration datasets including image super-resolution, image denoising and image compression artifact removal. We also conducted experiments to proof the scale-invariance of proposed SCN and show its advances over other multi-scale models.

{\small
\bibliographystyle{aaai}
\bibliography{ref}
}

\end{document}